\documentclass[journal]{IEEEtran}

\ifCLASSINFOpdf

\else

\fi

\usepackage{algorithm}
\usepackage{algorithmic}
\usepackage[]{hyperref}
\hypersetup{colorlinks=true,}
\usepackage{amssymb}
\usepackage{letltxmacro}
\usepackage{pifont}
\usepackage[cmex10]{amsmath}
\usepackage{algorithmic}
\usepackage{color}
\usepackage{cases}
\usepackage{bm}
\ifCLASSINFOpdf
\usepackage[pdftex]{graphicx}
\usepackage{epstopdf}
\else
\usepackage[dvips]{graphicx}
\fi
\ifCLASSOPTIONcompsoc
\usepackage[nocompress]{cite}
\else
\usepackage{cite}
\fi
\usepackage[caption = false]{subfig}
\usepackage{multirow}
\usepackage{filecontents,pgfplots}
\usepackage{tikz}
\usetikzlibrary{arrows,calc,shapes,snakes,positioning,matrix,arrows,decorations.pathmorphing,decorations.text}
\newcommand{\etal}{\textit{et al}.}

\begin{document}

\title{AutoIDS: Auto-encoder Based Method \\ 
	   for Intrusion Detection System}

\author{Mohammed~Gharib$^*$,~\IEEEmembership{Member,~IEEE,}                  
        Bahram~Mohammadi$^*$,%~\IEEEmembership{Fellow,~OSA,} 
        \\
        Shadi~Hejareh~Dastgerdi, %~\IEEEmembership{Member,~IEEE,}
        and~Mohammad~Sabokrou,~\IEEEmembership{Member,~IEEE}% <-this % stops a space
\thanks{$^*$ M. Gharib and B. Mohammadi contributed equally to this paper.}
%M. Shell was with the Department
%of Electrical and Computer Engineering, Georgia Institute of Technology, Atlanta,
\thanks{M. Gharib and M. Sabokrou are with the Institute for Research in Fundamental Sciences (IPM) (email: gharib, sabokro@ipm.ir).}% <-this % stops a space
\thanks{B. Mohammadi is with the Computer Engineering Department, Sharif University of Technology (email: bmohammadi@alum.sharif.edu).}% <-this % stops a
\thanks{S. H. Dastgerdi is with the Computer Engineering Department, Iran University of Science and Technology (email: shadi\_hejareh@iust.ac.ir).}% <-this % stops a
}


\maketitle

\begin{abstract}
 Intrusion Detection System (IDS) is one of the most effective solutions for providing primary security services. IDSs are generally working based on attack signatures or by detecting anomalies. In this paper, we have presented AutoIDS, a novel yet efficient solution for IDS, based on a semi-supervised machine learning technique. AutoIDS can distinguish abnormal packet flows from normal ones by taking advantage of cascading two efficient detectors. These detectors are two encoder-decoder neural networks that are forced to provide a compressed and a sparse representation from the normal flows. In the test phase, failing these neural networks on providing compressed or sparse representation from an incoming packet flow, means such flow does not comply with the normal traffic and thus it is considered as an intrusion. For lowering the computational cost along with preserving the accuracy, a large number of flows are just processed by the first detector. In fact, the second detector is only used for difficult samples which the first detector is not confident about them. We have evaluated AutoIDS on the NSL-KDD benchmark as a widely-used and well-known dataset. The accuracy of AutoIDS is 90.17\% showing its superiority compared to the other state-of-the-art methods. 
\end{abstract}


\IEEEpeerreviewmaketitle

\section{Introduction}
\label{sec:introduction}
	
\IEEEPARstart{N}{owadays}, providing security services in different computer networks is an issue of paramount significance. The principal security services required by almost all of the communication networks, irrespective of their types, are confidentiality, authenticity, non-repudiation, integrity, and availability. Cryptography is an effective solution for providing security services \cite{security_service}, however, it is ineffective against availability attacks. IDS is an alternative solution to provide availability as a crucial security service. Hence, IDS and cryptography complement each other to cover all of the aforementioned security services.

Intrusion detection methods can be generally categorized into two groups: (1) signatures based, and (2) anomaly detection based \cite{signature_anomaly}. The signature based methods can effectively detect network attacks which have been already known and their pattern is available, while they fail in detecting unknown and also zero-day intrusions. The latter solution, i.e., anomaly detection based, seems to be more effective due to its capability of detecting unknown and new attacks. Machine learning is one of the most promising techniques for anomaly detection \cite{ml}. Furthermore, by the recent exponential growth in the volume of the computer network data, this technique attracts even much more attention.
Machine learning technique basically falls into three categories, supervised, semi-supervised and unsupervised. 
	
The majority of the previously proposed methods for IDS are based on supervised learning \cite{supervised} meaning that they have access to both normal and abnormal samples in training duration and they are able to efficiently learn the characteristics of merely the available normal and abnormal flows. Accordingly, such methods can accurately detect abnormal packet flows only if they are available in training procedure (or be similar to training samples), otherwise, they fail to properly detect anomalies. Broadly, the generalization of supervised methods for the IDS task is not enough to be able to detect new types of attacks. Training an intrusion detection method in a semi-supervised fashion, i.e., training a model just on the normal traffic, without knowing anything about the abnormalities, is more general especially for new (unseen) network attacks.  Handling this problem leads to a generalized method that is able to act properly against new types of intrusions. 

\begin{figure}[t]
		\centering
		\includegraphics[width=0.41\columnwidth]{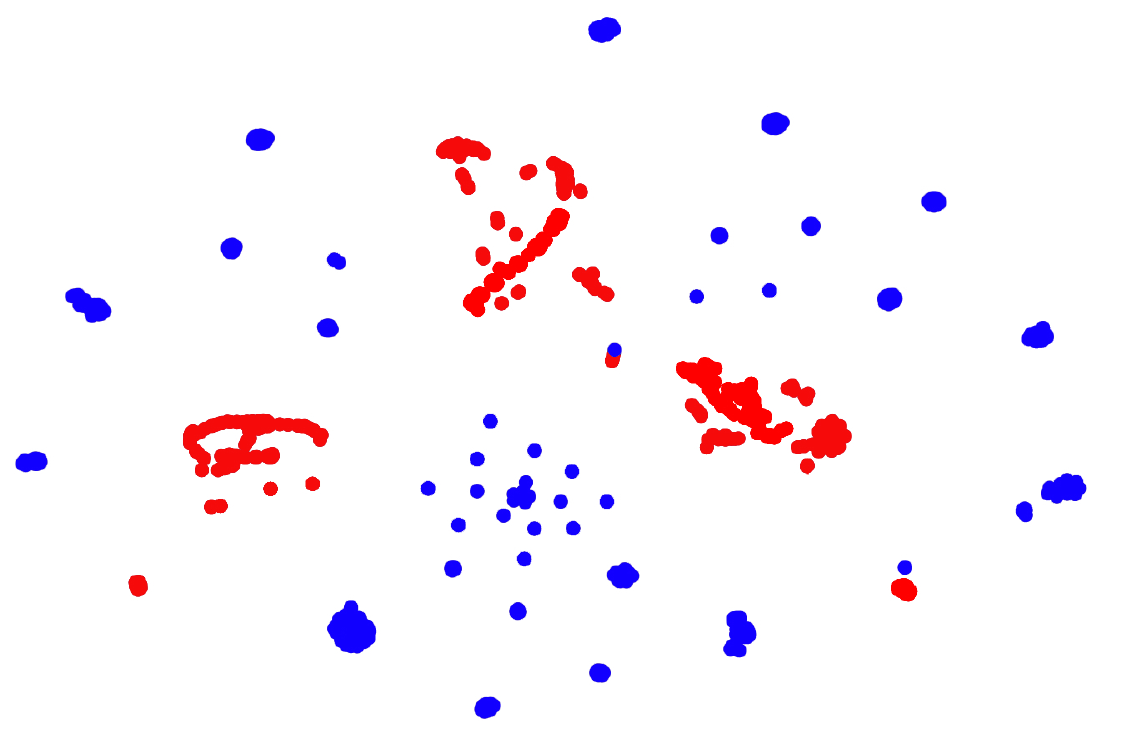}
			~~~~~~~	\includegraphics[width=0.41\columnwidth]{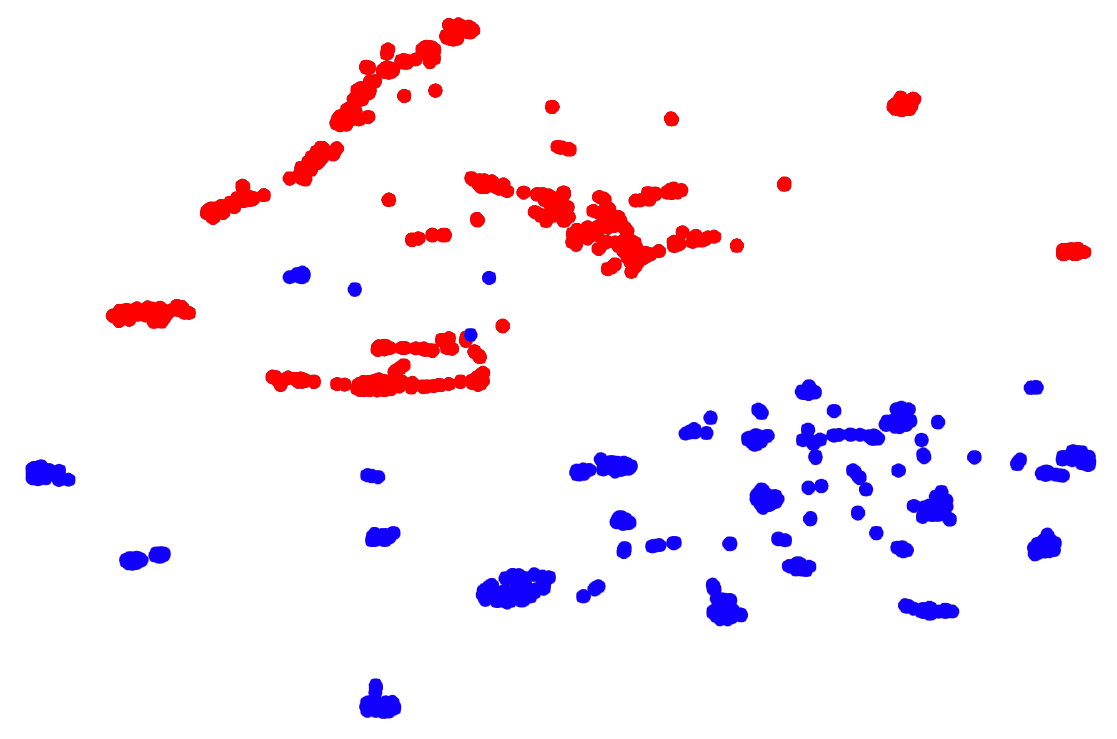}
		\caption{Visualizing the original normal (blue) and anomalous (red) packet flows using t-SNE before and after representing by the sparse AE. ({\it Left}:) original samples, ({\it Right}:) latent space of the sparse AE. As can be seen in this figure, the sparse AE makes the original normal and abnormal packet flows more separable from each other.}
			\label{fig:sparse-after}
\end{figure}
	
In this paper, we have proposed AutoIDS, a semi-supervised deep learning  method to precisely detect the anomalous traffic. We have named it AutoIDS, since it exploits two Auto-Encoders (AEs) for detecting anomalies in communication networks in a cascade manner. The key idea is to make a decision about the types of incoming network flows in two separate phases, each phase with a different detector. In this case, incoming network traffic is investigated in different feature spaces enabling us to differentiate the process of making decision about simply detectable and more complex incoming packet flows. Thus, the accuracy of the detection process is increased. Firstly, the sparse AE is exploited to detect the simple detectable flows. Then, the rest of packet flows are sent to a non-sparse AE. Thereby, distinguishing normal flows from complex anomalies are performed by AE with more accurate processing. From now on, for simplicity, we use AE instead of non-sparse AE, in short.  Cascading several weak detectors and early rejection of the most samples for improving the complexity and accuracy are widely used in machine learning community \cite{viola2001rapid,polikar2012ensemble}, especially for anomaly detection in videos \cite{sabokrou2018deep,sabokrou2017deep,sabokrou2017fast}. Inspired by such researches, we have devised an effectual method for IDS.
 
In computer networks, the incoming network flows are passing through a gateway. Therefore, they can be monitored at this point. Since such network traffic is heavy, their processing must be performed as fast as possible. To this end, a considerable volume of incoming flows analyzed and detected in the first step, i.e., by the sparse AE, with low time complexity. The rest of the incoming network flows are then sent to the AE with a more complicated decision making process. 
Note that the number of packet flows forwarded to the second phase, form a portion of the entire incoming traffic. 
Consequently, in addition to increasing the accuracy of anomaly detection, the time complexity is reduced noticeably. 

We have interestingly investigated that the sparsity value of a represented packet flow by the sparse AE as well as the calculated reconstruction error for an incoming flow through the AE which are trained merely on normal flows, are considered as two effective discriminative measures for hunting the abnormal traffic in computer networks. Proposing an efficient method to specify thresholds for both AE and sparse AE to distinguish between normal and anomalous traffic, is of crucial importance and accordingly forms part of our contribution. Additionally, to show the effectiveness of AutoIDS, we have investigated the power of the sparse AE to separate normal and abnormal flows using t-distributed Stochastic NEighbor (t-SNE) \cite{t-SNE}. In fact, t-SNE is an algorithm for dimensionality reduction which is appropriate for visualizing high dimensional data. As an instance and by utilizing t-NSE, Fig. \ref{fig:sparse-after}-left shows the feature vector of original normal and anomalous traffic, while Fig. \ref{fig:sparse-after}-right shows the represented original packet flows by the sparse AE, trained only on normal flows. In this  figure, there are 500 samples for each type of normal and abnormal traffic. As can be seen in Fig \ref{fig:sparse-after}, sparse representation of the packet flows is more separable than the original version of them. However, it is not sufficiently discriminative for separating difficult samples and just works well for simple ones. To cope with this weakness, as explained earlier, the decision about difficult samples is taken by the AE, which works based on the reconstruction error. 

The main contribution of this paper is proposing a novel, effective and accurate solution for detecting abnormal traffic with an acceptable time complexity. More specifically, our contributions are: (1) proposing a semi-supervised deep learning method to detect anomalies with higher performance, in terms of accuracy, compared to the other state-of-the-art solutions. To the best knowledge of us, this paper is one of the first deep learning based methods dealing with the intrusion detection as an semi-supervised approach, and (2) evaluating the proposed method under realistic circumstances and also showing its superiority. 

The rest of this paper is organized as follows. In Section \ref{sec:relatedWork}, we have briefly reviewed the other proposed approaches. A detailed explanation of AutoIDS has been provided in Section \ref{sec:proposedApproach}. Evaluation which includes experimental results alongside their analysis resides in Section \ref{sec:experimentalResult}. Finally, we have concluded this paper in Section \ref{sec:conclusion}.

	\begin{figure*}
		\includegraphics[width=\linewidth]{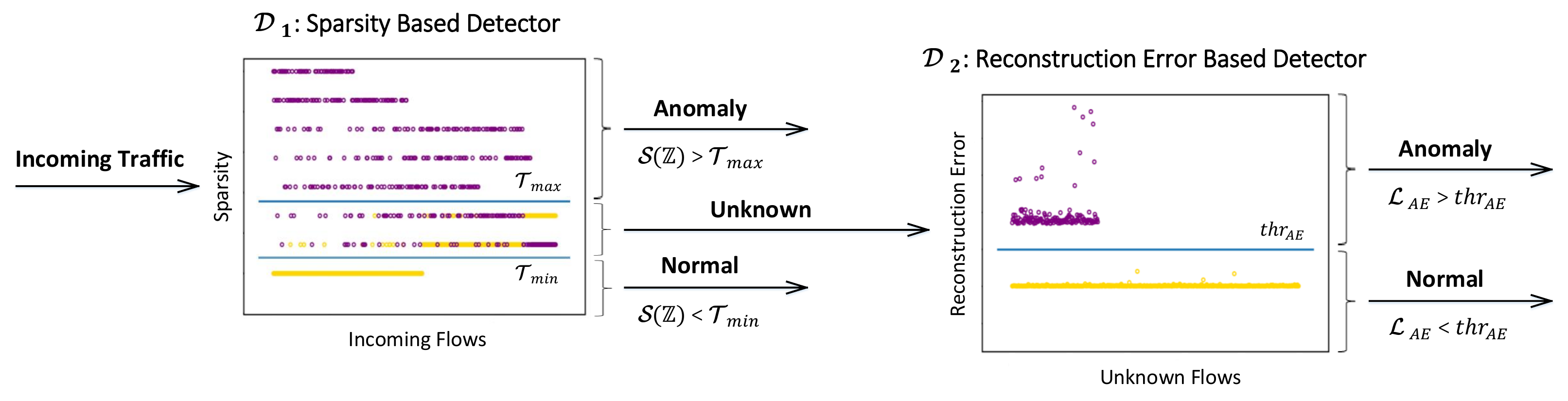}
		\caption{The process of anomaly detection in AutoIDS. Incoming packet flows are processed in two phases. More complex flows which the sparse AE is not confident about their types, are sent to the next phase. At this point, the AE makes a decision about the type of incoming traffic whether it is normal or not. For AE, accuracy is preferred to speed while this is opposite for sparse AE. In the first step, $ \tau_{min} $ and $ \tau_{max} $ are the minimum and maximum threshold for the sparse AE, respectively. Also, $ \mathcal{S}(\mathbb{Z}) $ shows the sparsity value. In the second step, $ \mathcal{L}_{\text{AE}} $ is the loss of AE which is calculated for each of incoming network flows and $ thr_{\text{AE}} $ is the threshold for separating the incoming traffic.}
		\label{fig:anomaly-detection}
	\end{figure*}

 

\section{Related Work}
\label{sec:relatedWork}
In this section, we have explored the intrusion detection method based on machine learning technique. Generally, IDSs can be divided into two groups: (1) signature based, and (2) anomaly detection based. In signature-based solutions, the incoming network traffic is compared with a database of signatures. Therefore, this method needs a continuous update in order to add new intrusion patterns. Although this way of approaching the problem is efficient for detecting the known attacks \cite{1}, it is incapable of detecting new (unseen) intrusions. On the other hands, anomaly detection based methods are able to detect the unknown and zero-day network attacks. In contrary to the signature based solutions, anomaly detection based methods, which are broadly based on machine learning technique, are more robust against unknown attacks\cite{2}. From the machine learning point of view, the anomaly detection based methods can be categorized into three approaches: (1) supervised, (2)semi-supervised, and (3) unsupervised \cite{3}. In supervised learning, sufficient amount of labeled data from both normal traffic and the known intrusions are fed to the neural network. In this case, the system can detect such network attacks more accurately. However, labeling all of the network flows is a highly time-consuming and also expensive process. Semi-supervised learning is used to utilize both labeled and unlabeled data to decrease the cumbersome of labeling packet flows, especially when the system faces a big volume of data and diverse abnormal samples. Unsupervised learning aims to cluster the samples with no knowledge. 
	
There are many algorithms based on supervised learning such as K-Nearest Neighbor (KNN) \cite{4}, neural network \cite{n1}, Support Vector Machine (SVM) \cite{6}, and etc. 
Chand \etal \cite{7} stacked a SVM, which is considered as an effective classifier for intrusion detection, followed by another ones such as 
BayesNet \cite{n2}, AdaBoost \cite{n1}, Logistic \cite{n3}, IBK \cite{n4}, J48 \cite{Quinlan1993}, RandomForest \cite{n6}, JRip \cite{Cohen1995}, and OneR \cite{Holte1993} to improve the detection accuracy. Tao \etal \cite{8} used the Genetic Algorithm (GA) alongside the SVM. Basically, GA can increase the true positive rate, decrease the error rate, and improve the classification time complexity of the SVM by selecting the best features. 
Ling \etal \cite{Ling2019} introduced a random forest feature selection algorithm and proposed a multi-classifier method. The random forest extracts optimal features for training SVM, decision tree, NaiveBayes \cite{domingos1997optimality}, and KNN classification algorithms. Then, deep learning is adopted to stack these classifiers. This proposed method improves the accuracy of intrusion detection in comparison with the majoring voting algorithms. 
Ashfaq \etal \cite{10}  proposed a fuzziness based semi-supervised learning approach to improve the performance of classifiers  for IDSs, however, it utilizes supervised learning. The proposed network has a single hidden layer, and the output is a fuzzy membership network. Fuzzy value is used for categorizing unlabeled samples. The classifier is retrained after incorporating each category separately into the original training set.
	
There are further literature works that have combined the supervised and unsupervised approaches to propose hybrid algorithms. Aljawarneh \etal \cite{12} developed a new hybrid model to minimize the time complexity of determining the feature association impact scale. This hybrid algorithm consists of several classifiers including J48, Meta Pagging \cite{Breiman1996}, RandomTree \cite{598994}, REPTree \cite{quinlan1987simplifying}, AdaBoostM1 \cite{Freund1996}, DecisionStump \cite{ref1}, and NaiveBayes.
	
Deep learning as one of the most popular branches of machine learning technique in the field of IDS, uses multiple hidden layers in order to extract features, automatically. However, machine learning based methods need an expert to manually extract features. Deep learning based solutions improve the detection accuracy compared to traditional methods\cite{d1}. 
Hodo \etal \cite{3} presented a taxonomy including shallow and deep neural networks for IDSs to demonstrate the effect of feature selection and feature extraction on the performance of the anomaly detection procedure. 
Javaid \etal\cite{sae_mlp} proposed a deep learning method based on sparse AE and softmax-regression to implement effective and flexible Network Intrusion Detection System (NIDS). This technique consists of two steps, (1) a sparse AE is used for feature learning due to its ease of implementation and good performance, and (2) softmax-regression is used for the classification purpose. 
Farahnakian \etal\cite{d3} proposed an approach based on a stacked AE. The whole neural network includes four AEs in which the output of each AE is considered as an input for the next AE, in the next layer. The stacked AE is followed by a softmax layer classifying incoming network flows into normal and anomaly. 
Shone \etal\cite{d4} proposed a Non-symmetric Deep AE (NDAE) for unsupervised feature learning and introduced the stacked NDAE and the random forest for classification. This model is a combination of deep and shallow learning to exploit their strengths and reduce the analytical overhead. 
Papamartzivanos \etal \cite{8620986} proposed a methodology that combines self-taught learning \cite{raina2007self} with MAPE-K framework \cite{kephart2003vision}. self-taught learning is utilized to transfer learning from unlabeled data to achieve high attack detection accuracy. Also, MAPE-K framework is used for delivering a self-adaptive IDS to identify the previously unseen intrusions via reconstructing of unlabeled data. 
The work of \cite{mohammadi2019end} inspired by \cite{sabokrou2018adversarially}, proposed a semi-supervised method based on deep generative method, where two deep neural networks are adversarially and jointly trained to generate abnormal flows and distinguish fake samples from the real ones.

\section{AutoIDS Algorithm }
\label{sec:proposedApproach}
In this section, we have introduced AutoIDS, the novel and effective approach to detect anomalies in different types of communication networks, with the aim of high accuracy and low computational cost. To cope with the supervised weaknesses, i.e., low generalization for detecting the unseen attacks and needed a numerous labeled samples, we have proposed an effective method with minimum supervision which is able to be efficiently trained by considering merely the normal samples. The key idea is to exploit a combination of an AE and a sparse AE in a cascade manner aiming to increase the accuracy and decrease the time complexity, at the same time.

AutoIDS consists of two detectors, $\mathcal{D}_1$ and $\mathcal{D}_2$ including sparse AE and AE, respectively. In $\mathcal{D}_1$, the criterion for distinguishing normal from anomalies is the sparsity. In fact, the sparse AE is trained only on normal traffic. Accordingly, it is optimized to provide a sparse representation from normal packet flows. Most probably anomalous traffic do not satisfy the imposed constraint and the sparse AE needs more active neurons to reconstruct the input, i.e., the sparsity value of their latent representations is not lower than a specific threshold. To reduce the overall complexity of AutoIDS, we have considered $\mathcal{D}_1$ as a lightweight detector, and it is allowed to work with high False Positive Rate (FPR). Flows which $\mathcal{D}_1$ are not certain about their types are sent to the next more accurate detector. 
	
In $\mathcal{D}_2$, reconstruction error is used as the criterion for separating different types of flows. Analogous with the sparse AE, the AE is also trained just using normal traffic. Hence, it learns to minimize the reconstruction error. Accordingly, when the reconstruction error of an incoming flow is more than a predefined threshold, it does not satisfy the desired constraint and thus it is considered as an anomaly. It is worth mentioning that, considering two different detectors, i.e., $\mathcal{D}_1$ and $\mathcal{D}_2$, empowers the system to make a decision about different types of traffic in two individual feature spaces. In this way, AutoIDS is able to differentiate normal traffic form anomalies with more accuracy.  Fig. \ref{fig:anomaly-detection} outlines our proposed solution for anomaly detection in computer networks. 
	
We have divided this section into three parts. The classification manner employed by the sparse AE has been explained in the first part. Then, we have presented a detailed explanation about separating normal and anomalous traffic by the AE in the second part. Finally, AE and sparse AE are considered as a whole and the way of detecting anomalies by AutoIDS has been investigated. 
	 
\subsection{$\mathcal{D}_1:$ Sparsity Based Detector}
\begin{figure}
		\includegraphics[width=\linewidth]{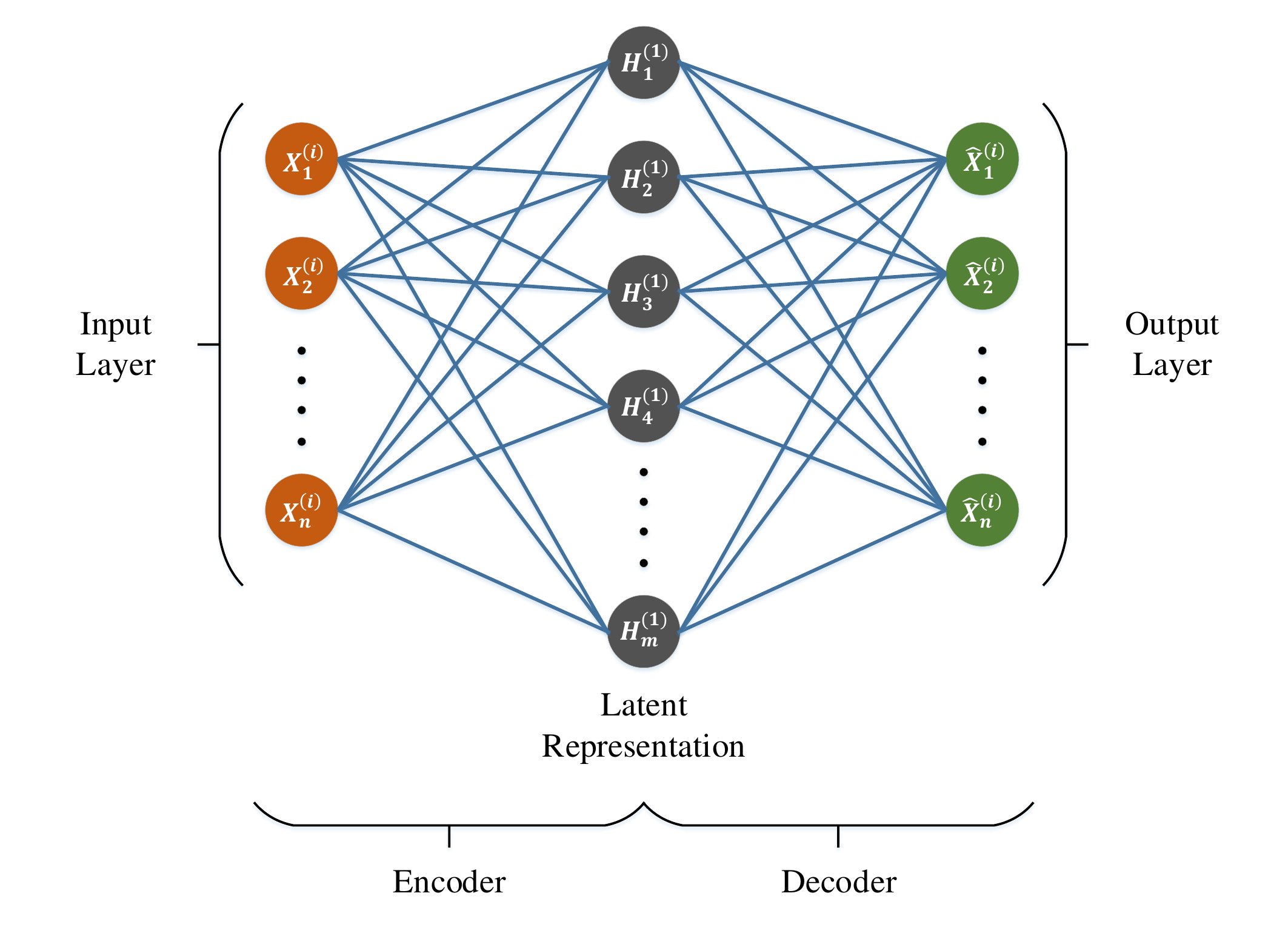}
		\caption{The overall scheme of our proposed sparse AE. In this figure, $X^{(i)}_{j}$ and $X'^{(i)}_{j}$ are the $j-th$ feature of $i-th$ sample and its reconstructed version, respectively. $H^{(i)}_{j}$ shows the $j-th$ feature of the latent representation of $i-th$ sample. In AutoIDS $n = 122$ and $m = 140$.}
		\label{fig:sparse-scheme}
	\end{figure}

Sparse AEs are originally used for unsupervised feature learning while they improve generality \cite{kavukcuoglu2010learning}. However, we have exploited the value of the sparsity in the latent representation like what has been also considered as an effective solution in \cite{sabokrou2016video}. The overall scheme of the sparse AE, which has been used in our proposed method, is depicted in Fig. \ref{fig:sparse-scheme}. As it is clear in this figure, the sparse AE consists of two major parts, encoder and decoder. Furthermore, to the latent representation becomes over-complete, the number of neurons in the latent representation (m) should be considered more than the size of the input layer (n), i.e., $ m>n $.  The applied constraint on the sparse AE during the training phase, is sparsity. As explained previously, the sparse AE is merely trained on normal traffic. Therefore, the sparse AE parameters ($ \theta_{\text{SAE}} $) are optimized based the normal flows. The proposed spares AE for detecting the anomalous packet flows is trained based on Equation \ref{eq:loss-sae}.
	
	\begin{equation} 
	\begin{split}
	\begin{array}{c@{\qquad}c}
	\label{eq:loss-sae}
	\mathcal{L}_\text{SAE} = \|X - \hat{X} \|^{2}_{2} + \sum_{j=1}^{m} KL(\rho \| \hat{\rho}_j) \\ \\
	\hat{X} = \sigma_2(W_2\sigma_1(W_1X+b_1)+b_2)
	\end{array}
	\end{split} 
	\end{equation}
	
where $ X \in \mathbb{R}^n $ is the characteristic set of each incoming flow, $ \hat{X} \in \mathbb{R}^n $ is the characteristic set of the reconstructed version of each incoming flow, $ \sigma_1 $ and $ \sigma_2 $ are activation functions, $ W_1 $ and $ W_2 $ are encoder and decoder weight matrices, respectively, $ b_1 $ and $ b_2 $ are bias terms, $ \rho $ is the sparsity parameter, $ \hat{\rho}_j $ is the expected value for $ j-th $ unit in hidden layer and $ W_1\times X = \mathbb{Z} \in \mathbb{R}^m$ is the latent representation.
	
With respect to Equation \ref{eq:loss-sae}, the sparse AE is trained to minimize $ \mathcal{L}_{\text{SAE}} $ for normal flows and also produce $\mathbb{Z} $ whose sparsity value, i.e., $ \mathcal{S}(\mathbb{Z}) $, is lower than $ \rho $, where $\mathcal{S}(\mathbb{Z}=\mathcal{D}_1(X))$ is the sparsity value of the representation of X using $\mathcal{D}_1$ network and can be calculated based on the Equation \ref{eq:sv}. 

    \begin{equation}
        \mathcal{S}(\mathbb{Z})=\frac{\#_{\ne0}(\mathbb{Z})}{m}
        \label{eq:sv}
    \end{equation}

Where $\#_{\ne0}(\mathbb{Z})$ and $m$ determine the total number of active neurons (non-zero elements) and total number of neurons (including active and inactive neurons) in $\mathbb{Z}$, respectively.
As can be seen in Fig. \ref{fig:anomaly-detection}, to speed up the decision making process of AutoIDS about the incoming traffic, two thresholds are determined in this phase, $ \tau_{min} $ and $ \tau_{max} $. Flows with $ \mathcal{S}(\mathbb{Z}) > \tau_{max} $ are detected as anomaly. On the contrary, when an incoming flow follows the concept of normal traffic and its sparsity value is lower than $ \tau_{min} $, it is detected as normal. Otherwise, i.e., $\tau_{min} < \mathcal{S}(\mathbb{Z}) < \tau_{max} $, the incoming packet flow is considered as an unknown flow and thus it needs further process due to lack of precision in $ \mathcal{D}_1 $. Although anomalies are detected by $ \mathcal{D}_1 $ are not complicated, including both normal and anomalous samples, their number is noticeable. In fact, a large volume of incoming traffic are detected by $ \mathcal{D}_1 $ with low time complexity.

\subsection{$\mathcal{D}_2:$ Reconstruction Error Based Detector}
\begin{figure}
		\includegraphics[width=\linewidth]{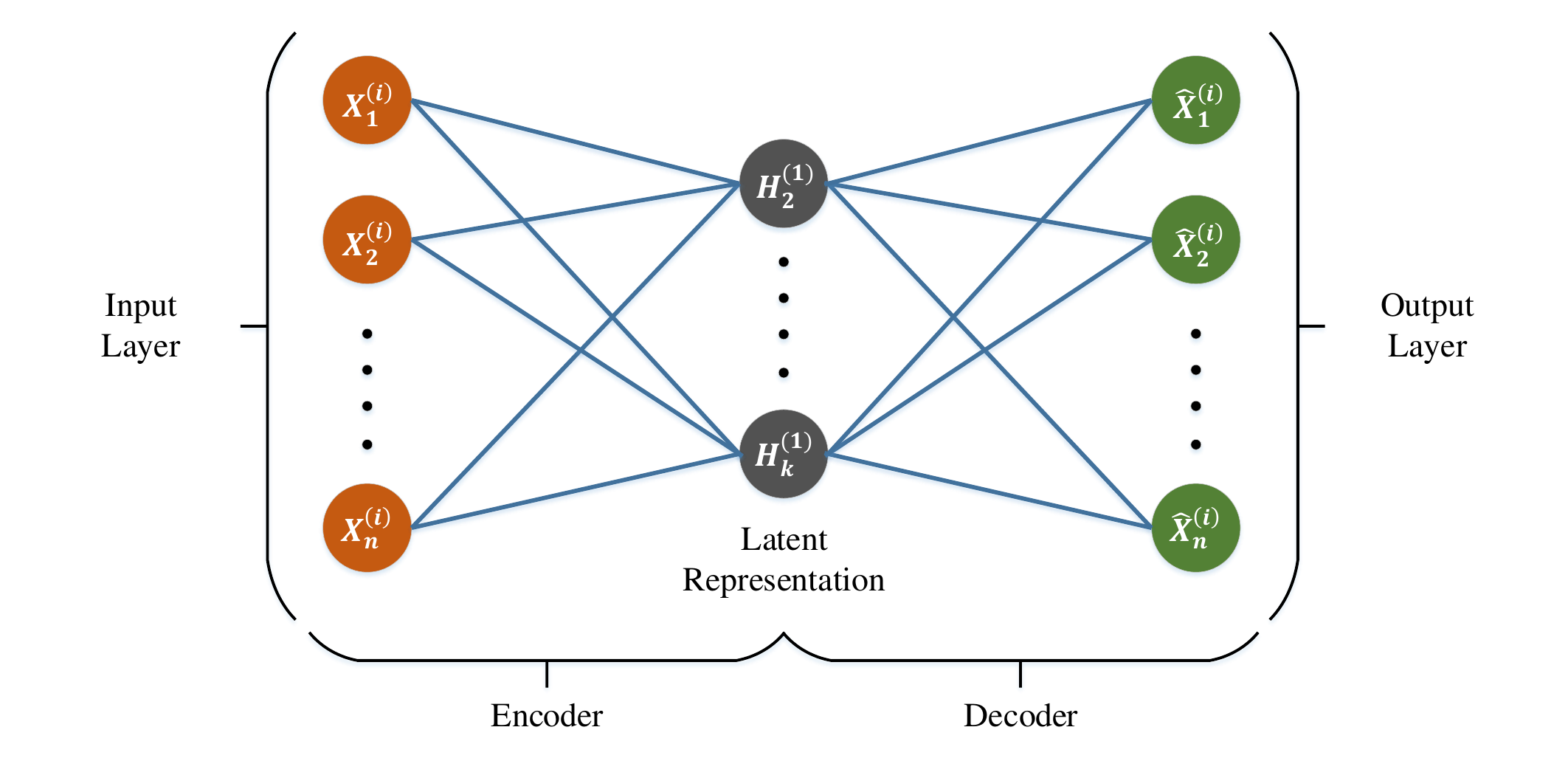}
		\caption{The overall scheme of our proposed AE. In this figure, $X^{(i)}_{j}$ and $X'^{(i)}_{j}$ are the $j-th$ feature of $i-th$ sample and its reconstructed version, respectively. $H^{(i)}_{j}$ shows the $j-th$ feature of the latent representation of $i-th$ sample. In AutoIDS $n = 122$ and $k = 80$.}
		\label{fig:auto-scheme}
	\end{figure}

AEs are mainly used for dimensionality reduction and unsupervised feature learning \cite{kavukcuoglu2010learning}. However, as has also been investigated in \cite{sabokrou2018avid}, leveraging the reconstruction error is very useful. The overall architecture of our desired AE is represented in Fig. \ref{fig:auto-scheme}. Unlike the sparse AE, the whole structure of the AE is used to reconstruct the incoming flows. In $ \mathcal{D}_2 $, the AE is forced to reconstruct only the normal flows using a hidden layer with the lower number of neurons (k) compared to the number of input neurons (n) in the training phase ($ k<n $). The loss function of AE is shown in Equation \ref{eq:loss-ae} called squared euclidean error. Note that the loss function of AE is similar to sparse AE, but the sparsity constraint no longer exists.
	
	\begin{equation} 
	\begin{split}
	\begin{array}{c@{\qquad}c}
	\label{eq:loss-ae}
	\mathcal{L}_{\text{AE}} = \|X - \hat{X} \|^{2}_{2}
	\end{array}
	\end{split} 
	\end{equation}
	
$ \mathcal{L}_{AE} $ is minimized for normal traffic and $ \theta_{\text{AE}} $, i.e., the parameters vector of AE, is optimized on reconstructing the incoming normal flows with low reconstruction error. In fact, we have aimed to exploit the lack of ability of the AE to reconstruct incoming abnormal flows  to distinguish different types of traffic. Reconstruction error is calculated for all of the incoming packet flows entering the AE. If it be lower than a predetermined threshold ($ thr_{\text{AE}} $), the packet flow is considered as normal, otherwise, it is detected as anomaly.
	
\subsection{Training AutoIDS}
In AutoIDS, $ \mathcal{D}_{1} $ and $ \mathcal{D}_{2} $ learn the concept of normal traffic, individually. In other words, both of them are trained on all of the normal packet flows included by the training set and thus their learning process is not done according Fig. \ref{fig:anomaly-detection}. $ \mathcal{D}_{1} $ learns to provide a sparse representation from normal traffic. Consequently, the number of active neurons in the hidden layer should increase in case of facing anomalous samples. Also, $ \mathcal{D}_{2} $ is trained to provide a compressed representation of normal flows. When an abnormal packet flow enters $ \mathcal{D}_{2} $, it is not capable of properly providing the reconstructed version. Therefore, the reconstruction error increases which is the criterion for distinguishing different types of incoming traffic from each other.

\subsection{Anomaly Detection}
In this section, we have cohesively described the manner of detecting anomalies in AutoIDS. Suppose an incoming flow enters the computer network. At first, as it is shown in Fig. \ref{fig:anomaly-detection}, the incoming flow enters the sparse AE. Formal representation of processing the incoming flow in the sparse AE is given as follows.
	
	\begin{equation}\nonumber
	\begin{split}
	\begin{array}{c@{\qquad}c}
	\label{eq:sparse-formal}
	\mathcal{D}_1(x)=\begin{cases}
	Normal, & \text{if $ \mathcal{S}(\mathbb{Z})<\tau_{min} $}.\\
	Anomaly, & \text{if $ \mathcal{S}(\mathbb{Z})>\tau_{max} $}.\\
	Unknown, & \text{otherwise}.
	\end{cases}
	\end{array}
	\end{split} 
	\end{equation}
	
A large amount of normal and anomalous flows, which are not complicated, are detected in this phase. Accordingly, Flows which their behavior are more complex, i.e., unknown flows, are send to the next detector for further processing irrespective of their actual types. This cascading method results in increasing accuracy, while the time complexity is reduced compared to each detector individually. The second phase is shown in below.
	
	\begin{equation}\nonumber
	\begin{split}
	\begin{array}{c@{\qquad}c}
	\label{eq:auto-formal}
	\mathcal{D}_2(x \in Unknown)=\begin{cases}
	Anomaly, & \text{if $ \mathcal{L}_{\text{AE}}>thr_{\text{AE}} $}.\\
	Normal, & \text{otherwise}.
	\end{cases}
	\end{array}
	\end{split} 
	\end{equation}
	
The second phase is more complex and more precise. Hence, it is suitable for distinguishing anomalies from normal traffic in case that they are very similar to each other. The point should be noticed is the way of determining $ thr_{\text{AE}} $. In fact, this threshold is specified based on Receiver Operating Characteristic (ROC) curve. The best threshold is the one where $ recall $ is equal (roughly equal) to $ 1-fpr $. If we consider $ thr_{AE} $ as the best threshold, then: 
	
	\begin{equation}\nonumber
	\begin{split}
	\begin{array}{c@{\qquad}c}
	\label{eq:auto-formal}
	\text{Recall}(thr_{\text{AE}}) \simeq 1-\text{FPR}(thr_{\text{AE}})
	\end{array}
	\end{split} 
	\end{equation}

\section{Evaluation}
\label{sec:experimentalResult}
In this section, We have evaluated the performance of our proposed method, i.e., AutoIDS, on NSL-KDD\footnote{Available at https://www.unb.ca/cic/datasets/nsl.html} dataset \cite{nsl}. We have compared the results of AutoIDS with the state-of-the-art intrusion detection methods which are based on anomaly detection. We have also expressed a comprehensive analysis of the obtained results confirming that our method is able to be performed as an effective method for intrusion detection. 
	
\subsection{Dataset}
\label{sec:dataset}
KDDCUP'99 dataset is constructed based on DARPA'98 IDS evaluation program \cite{darpa}. This dataset is provided by Stolfo \etal \cite{stolfo}. A large number of duplicate records is the fundamental problem of this dataset. Therefore, classifiers had a bias toward the frequent records. Tavallaee \etal \cite{nsl} improved KDDCUP'99 by removing the duplicate records, and  named  it NSL-KDD.  Furthermore, the reasonable number of samples in both training and test sets has enabled researchers to use the entire dataset for the process of evaluation and not to select them randomly. Thus, they are able to fairly compare their results with other proposed methods. Each of flows in NSL-KDD includes 41 features. Three of these features are non-numeric values and the processing of such features is not possible. To solve this issue, a pre-processing for converting those three features to numerical type is needed (See subsection \ref{sec:preprocessing}). 
	
NSL-KDD is composed of two subsets: (1) KDDTrain$ ^{+} $ and, (2) KDDTest$^{+} $. The available attack categories in this dataset are (1) Denial of Service (DoS), (2) probing, (3) User to Root (U2R), and (4) Root to Local (R2L).  KDDTrain$^{+} $ contains 24 attack types belonging to the above-mentioned abnormal classes. It is worth mentioning that, there are 17 attack types in  KDDTest$^{+} $ which is not included by KDDTrain$^{+} $. In other words, the test process is performed on some malicious flows (network intrusions) which certainly has not been got involved in the training procedure. This characteristic leads to examining the generalization of different proposed approaches that use NSL-KDD for the evaluation process. Table \ref{tab:nslkdd-spec} shows the detailed information of the NSL-KDD dataset.
	
	\begin{table}
		\centering
		\renewcommand{\arraystretch}{1.5}
		\caption{The specification of the NSL-KDD dataset.}
		\label{tab:nslkdd-spec}
		\begin{tabular}{|c||c|c|c|}
			\cline{3-4}
			\multicolumn{2}{c|}{} & \multicolumn{2}{c|}{\textbf{Number of Records}} \\
			\hline
			\textbf{Category} & \textbf{Class} & \textbf{ KDDTrain$^{+} $} & \textbf{ KDDTest$^{+} $} \\
			\noalign{\hrule height 1.5pt}
			\multirow{4}{*}{\textbf{Anomaly}} & DoS & 45927 & 7458 \\ \cline{2-4}
			& Probing & 11656 & 2421 \\ \cline{2-4}
			& R2L & 995 & 2754 \\ \cline{2-4}
			& U2R & 52 & 200 \\ \noalign{\hrule height 1.5pt}
			\textbf{Normal} & --- & 67343 & 9711 \\
			\noalign{\hrule height 1.5pt}
			\textbf{Total} & --- & 125973 & 22544 \\
			\hline
		\end{tabular}
	\end{table}
	
\subsection{Pre-processing}
\label{sec:preprocessing}
As stated previously, since NSL-KDD contains non-numerical values (protocol, service and flag), a prepossessing is essential to process the entering flows. We have converted these three features to numerical values using one-hot encoding. For instance, three protocols are encoded as follows.

	\begin{equation}\nonumber
	\begin{split}
	\begin{array}{c@{\qquad}c}
	\label{eq:encoded-nslkdd}
	\text{one-hot encoded}=\begin{cases}
	(0,0,1), & \text{if~the~protocol~is~tcp}.\\
	(0,1,0), & \text{if~the~protocol~is~udp}.\\
	(1,0,0), & \text{if~the~protocol~is~icmp}.
	\end{cases}
	\end{array}
	\end{split} 
	\end{equation}
	
Services and flags are encoded in the same way. There are three different protocols, 70 different services, 11 different flags and 38 other numeric features. Accordingly, there are 122 features in total. After converting all columns (features) to numerical values, standardization and also normalization is needed to be ready for the training phase. Features are standardized based on the Equation \ref{eq:standardization}.
	
	\begin{equation}
	\begin{split}
	\begin{array}{c@{\qquad}c}
	\label{eq:standardization}
	x(i) = \dfrac{x(i) - mean(x(i))}{standard\_deviation(x(i))} \\ \\
	\forall~i\in[1, 122]
	\end{array}
	\end{split} 
	\end{equation}
	
Where $ x(i) $ is the standardized form of $ i-th $ feature. Standardization procedure makes the mean and the standard deviation of each feature equal to 0 and 1, respectively. Then, normalization is done using the Equation \ref{eq:normalization}.

	\begin{equation}
	\begin{split}
	\begin{array}{c@{\qquad}c}
	\label{eq:normalization}
	x(i) = \dfrac{x(i) - min(x(i))}{max(x(i))-min(x(i))} \\ \\
	\forall~i \in[1, 122]
	\end{array}
	\end{split} 
	\end{equation}
	
Where $ x(i) $ is the normalized form of $ i-th $ feature. Finally, $ x(i) $ is fed to both encoder-decoder networks in the stage of training.

\subsection{Implementation Details}
We have implemented AutoIDS using Keras framework \footnote{https://keras.io/}.
The evaluations are performed on GOOGLE COLAB\footnote{https://colab.research.google.com}. The detailed information and learning parameters of $ \mathcal{D}_1 $  and $ \mathcal{D}_2 $ are provided in Table \ref{tab:sparse-spec}.
	
	\begin{table}
		\centering
		\renewcommand{\arraystretch}{1.5}
		\caption{AE and sparse AE specifications.}
		\label{tab:sparse-spec}
		\resizebox{1\textwidth}{!}{\begin{minipage}{\textwidth}
				\begin{tabular}{l|l|l}
				
					\multicolumn{1}{c}{Parameters} & \multicolumn{1}{c}{AE} & \multicolumn{1}{c}{Sparse AE} \\
					
					\hline
					
					Hidden layers & 1 & 1 \\ 
					Hidden layer size (neurons) & 80 & 140 \\
					Encoder activation function & ReLU & ReLU \\
					Decoder activation function & Sigmoid & Sigmoid \\
					Loss function & MSE & MSE \\
					Optimizer & Adam & Adam \\
					Regularizer & -- & 10e-5 \\
					
				\end{tabular}
		\end{minipage} }
	\end{table}
	
AutoIDS has been evaluated based on five measures, accuracy ($ ACC $), precision ($ PR $), recall ($ RE $), f-score ($ FS $) and false positive rate ($ FPR $). These measures have been calculated through the Equation \ref{eq:measures}.
	
	\begin{equation}
	\begin{split}
	\begin{array}{c@{\qquad}c}
	\label{eq:measures}
	ACC = \dfrac{TP+TN}{TP+FN+TN+FP} \\ \\
	PR = \dfrac{TP}{TP+FP} \\ \\
	RE = \dfrac{TP}{TP+FN} \\ \\
	FS = \dfrac{2*PR*RE}{PR+RE} \\ \\
	FPR = \dfrac{FP}{FN+FP} \\ \\
	\end{array}
	\end{split} 
	\end{equation}
	
Where True Positive (TP) and False Negative (FN) refers to those anomalous flows which are classified correctly as anomaly and mistakenly as normal, respectively. Similarly, the True Negative (TN) and False Positive (FP) are the normal flows which are classified as normal and abnormal, respectively. 
	
\subsection{Experimental Results}
We have compared our proposed method, AutoIDS, with the other state-of-the-art solutions. Also, the performance of AutoIDS against changing the anomalous traffic ratio to the normal traffic has been evaluated. Finally, We have investigated the performance of each detector individually, in the second part.
	
\textbf{Comparison with the previous methods:}
We have compared AutoIDS with methods reporting the result of 2-class performance and uses the same dataset for the test phase, i.e.,  KDDTest$^{+} $. Table \ref{tab:accuracy_comparison} shows that AutoIDS outperforms the state-of-the-art anomaly detection based methods for IDS, in terms of accuracy. Note that AutoIDS has a better performance even in comparison with the methods using supervised learning approach. 
	
	\begin{table}
		\centering
		\renewcommand{\arraystretch}{1.5}
		\caption{AutoIDS accuracy comparison when the training process is performed on KDDTrain$^{+}$ and the test procedure is done on KDDTest$^{+}$.}
		\label{tab:accuracy_comparison}
		\begin{tabular}{ccc}
			
			\hline
			
			Method & Supervised & $ Accuracy (\%) $ \\
			
			\noalign{\hrule height 1.5pt}
			
			RNN-IDS \cite{rnn_ids} & \checkmark & 83.28 \\	
			
			DCNN \cite{lstm_dcnn} & \checkmark & 85.00 \\
	
			Sparse AE and MLP \cite{sae_mlp} & \checkmark & 88.39 \\
			
			Random Tree \cite{randomTree} & \checkmark & 88.46 \\
	
			LSTM \cite{lstm_dcnn} & \checkmark & 89.00 \\
			
			Random Tree and NBTree \cite{randomTree} & \checkmark & 89.24 \\

			\hline \hline
			
	    	AE \cite{ae_dae} &  & 88.28 \\
			
			De-noising AE \cite{ae_dae} &  & 88.65 \\
			
			Ours (AutoIDS) &  & \underline{\textbf{90.17}} \\
			
			\hline
			
		\end{tabular}
	\end{table}
	
Regarding Table \ref{tab:accuracy_comparison} and the characteristics of NSL-KDD dataset,  AutoIDS is more efficient for detecting unseen network attacks. This superior performance has originated from the feature of generalization. To showcase the generality of AutoIDS, both training and test process are performed on KDDTrain$^{+}$. In this case, the test procedure is done on anomalies which are most probably available in the training set. Having learned all types of anomalous traffic, the neural network gets familiar with their concept and then distinguishes anomalies from normal flows more precisely. In this experiment, KDDTrain$ ^{+} $ is divide into three subsets, training set, validation set and test set. The data distribution of these subsets is represented in Table \ref{tab:data_distriburion}. Note that flows are selected randomly for each subset. 
	
	\begin{table}
		\centering
		\renewcommand{\arraystretch}{1.5}
		\caption{Data distribution of training, validation and test sets.}
		\label{tab:data_distriburion}
		\begin{tabular}{cccc}
			\cline{2-4}
			\multicolumn{1}{c}{} & \multicolumn{3}{c}{\textbf{ KDDTrain$^{+} $}} \\
			\hline
			\textbf{Class} & \textbf{Training} & \textbf{Validation} & \textbf{Test} \\
			\noalign{\hrule height 1.5pt}
			\textbf{Normal} & 53873 & 6735 & 6735 \\
			\hline
			\textbf{Anomaly} & --- & 6735 & 6735 \\
			\hline
		\end{tabular}
	\end{table}
	
Table \ref{tab:generalization} shows the accuracy of AutoIDS compared to the other state-of-the-art approaches in Table \ref{tab:accuracy_comparison} which also report the accuracy of their proposed solution on  KDDTrain$^{+}$. In fact, we need the other approaches evaluating their proposed solution using both KDDTrain$^{+} $ and KDDTest$^{+} $, otherwise, the generalization is not understood properly. Although AutoIDS has a better performance compared to the other semi-supervised approaches \cite{ae_dae}, it is less accurate than supervised ones. Supervised methods learn the concept of available intrusions in the training set. Obviously, when they face same attack types in the test phase, separating incoming flows is done more accurately. Table \ref{tab:accuracy_comparison} along with Table \ref{tab:generalization} confirms that AutoIDS  outperforms the other solutions against new and unknown attacks indicating the generalization of this method.
	
	\begin{table}[t]
	
		\renewcommand{\arraystretch}{1.5}
		\caption{AutoIDS performance comparison with the other state-of-the-art solutions when both training and test phases are performed on KDDTrain$^{+}$. Note that KDDTest$^{+}$ is not used.}
		\label{tab:generalization}
		\begin{tabular}{ccccc}
	
			\hline
			
			Method & $ ACC (\%) $ & $ PR (\%) $ & $ RE (\%) $ & $ FS (\%) $ \\
			
			\noalign{\hrule height 1.5pt}
			
			AE \cite{ae_dae} & 93.62 & 91.39 & 96.33 & 93.80 \\
			
			De-noising AE \cite{ae_dae} & 94.35 & 94.26 & 94.43 & 94.35 \\
			
			Ours (AutoIDS) \cite{ae_dae} & 96.45 & \underline{\textbf{95.56}} & \underline{\textbf{97.43}} & 96.49 \\

			Sparse AE and MLP \cite{sae_mlp} & 98.30 & --- & --- & \underline{\textbf{98.84}} \\
			
			RNN-IDS \cite{rnn_ids}  & \underline{\textbf{98.81}} & --- & --- & --- \\
			
			\hline
			
		\end{tabular}
	\end{table}
	
\begin{figure}[t]
\begin{center}
\begin{tikzpicture}
  \begin{axis}[width=9.5cm, height=6cm,
    symbolic x coords = {10, 15,20, 25,30,35, 40,45, 50},
    legend pos = south east,
    xlabel={Percentage of outliers (\%)},
    ylabel={$F$-Score},
    y label style={at={(axis description cs:0.05,.5)}},
  ]
  \addplot+[smooth,red] coordinates { (10,0.58)(20,0.73)(30,0.79)(40,0.82)(50,0.85)};
  \addplot+[smooth,cyan] coordinates { (10,0.55)(20,0.70)(30,0.76)(40,0.80)(50,0.83)};
  \addplot+[smooth,green] coordinates { (10,0.54)(20,0.68)(30,0.73)(40,0.77)(50,0.80)};

  \legend{{\footnotesize Ours (AutoIDS)}, {\footnotesize De-nosing AE\cite{ae_dae}}, {\footnotesize AE\cite{ae_dae}}}
  \end{axis}
\end{tikzpicture}
\end{center}
   \caption{Comparisons of f-scores for different percentages of anomalous flows involved in the experiment.}
		\label{fig:anomaly-amount-fs}
\end{figure}
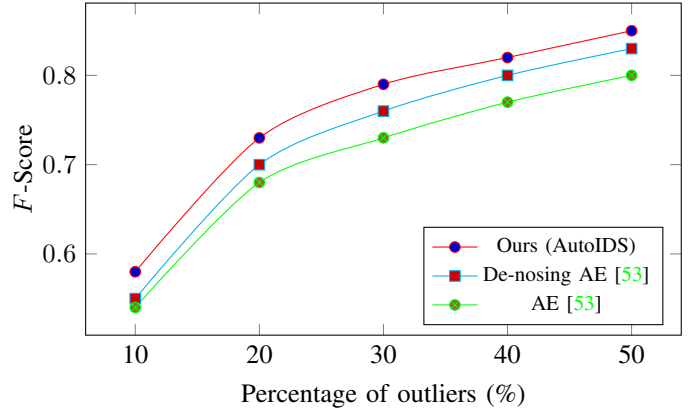

Generally, in the real world computer networks, abnormal traffic constitutes a lower portion of total traffic. Hereupon, we have evaluated AutoIDS using the f-score measure with different test datasets when NSL-KDD is used. The amount of anomalies is considered from 10 percent of the normal to 50 percent, with the step of 10. Fig. \ref{fig:anomaly-amount-fs} shows that AutoIDS outperforms the other two semi-supervised solutions in \cite{ae_dae}.

\textbf{Ablation Study:}
Both $ \mathcal{D}_1 $ and $ \mathcal{D}_2 $ can be used for anomaly detection separately, but exploiting both of them results in gaining more accuracy. Table \ref{tab:final-results} shows the performance of AutoIDS when $ \mathcal{D}_1 $ and $ \mathcal{D}_2 $ work based on sparsity ($ \mathcal{S} $)  and reconstruction error ($ \mathcal{R} $)  respectively, i.e., $ \mathcal{D}_1^{[\mathcal{S}]} \rightarrow \mathcal{D}_2^{[\mathcal{R}]} $. We have also reported the performance of each detector individually as an efficient detector when they both work based on reconstruction error, i.e., $ \mathcal{D}_1^{[\mathcal{R}]} $ and $ \mathcal{D}_2^{[\mathcal{R}]} $.
	
	\begin{table}
		\centering
		\renewcommand{\arraystretch}{2}
		\caption{AutoIDS Performance comparison with each detector individually when the training process is performed on KDDTrain$^{+}$ and the test procedure is done on KDDTest$^{+}$}
		\label{tab:final-results}
		\begin{tabular}{l c c c}
	
			\hline
			\hline
			& AutoIDS (\textbf{$ \mathcal{D}_1^{[\mathcal{S}]} \rightarrow \mathcal{D}_2^{[\mathcal{R}]} $}) & \textbf{$ \mathcal{D}_2^{[\mathcal{R}]} $} & \textbf{$ \mathcal{D}_1^{[\mathcal{R}]} $} \\ \cline{2-4}
			
			$ ACC $ ($ \% $) & \underline{\textbf{90.17}} & 89.14 & 87.50 \\ 
			
			$ PR $ ($ \% $) & 90.80 & 91.31 & \underline{\textbf{91.72}} \\ 
			
			$ RE $ ($ \% $) & \underline{\textbf{92.05}} & 89.43 & 85.79 \\ 
			
			$ FS $ ($ \% $) & \underline{\textbf{91.42}} & 90.36 & 88.65 \\ 
			
			$ FPR $ ($ \% $) & 12.33 & 11.25 & \underline{\textbf{10.24}} \\
			
			\hline
			\hline
	
		\end{tabular}
	\end{table}
	
	\begin{table}[t]
		\centering
		\renewcommand{\arraystretch}{2}
		\caption{Average test time comparison per each incoming flow for AutoIDS and each of detectors individually when the training process is performed on KDDTrain$^{+}$ and the test procedure is done on KDDTest$^{+}$}
		\label{tab:time-comparison}
		\begin{tabular}{c c}
		
			\hline
			\hline
			
			\textbf{AutoIDS ($ \mathcal{D}_1^{[\mathcal{S}]} \rightarrow \mathcal{D}_2^{[\mathcal{R}]} $)} & \underline{\textbf{137}} $ \mu sec $ \\
			
			\textbf{$ \mathcal{D}_2^{[\mathcal{R}]} $} & 195 $ \mu sec $ \\
			
			\textbf{$ \mathcal{D}_1^{[\mathcal{R}]} $} & 202 $ \mu sec $ \\
			
			\hline
			\hline
		
		\end{tabular}
	\end{table}	
	
Table \ref{tab:final-results} shows that AutoIDS ($ \mathcal{D}_1^{[\mathcal{S}]} \rightarrow \mathcal{D}_2^{[\mathcal{R}]} $) outperforms $ \mathcal{D}_1^{[\mathcal{R}]} $ and $ \mathcal{D}_2^{[\mathcal{R}]} $ in terms of accuracy, recall and f-score. AutoIDS processes incoming traffic in two steps. We have concentrate on simple normal and anomalies in the step one, while the decision about more complex traffic is made in the next step. Hence, AutoIDS performs more precise compared to each detector separately.

The computational cost for processing  a packet flow is also taken into account due to its significance for proposing an appropriate solution. In fact, the time of processing an incoming flow should not be the bottleneck as much as possible. Once again we have compared these approaches, but this time in terms of time complexity in the test phase. 
Table  \ref{tab:time-comparison} confirms that AutoIDS, i.e., $ \mathcal{D}_1^{[\mathcal{S}]} \rightarrow \mathcal{D}_2^{[\mathcal{R}]} $, can effectively detect anomalies and it is faster than each of detectors by a considerable margin. Since in AutoIDS all of the incoming flows are just encoded by $ \mathcal{D}_1 $ and also a large amount of normal and abnormal traffic are detected in the first step, our proposed method achieve lower time complexity in comparison with the other ones. In fact, all of incoming flows are encoded and then decoded when each detector is used individually and this way of approaching the problem makes them slower.

Table.~\ref{tab:final-results} and Table.~\ref{tab:time-comparison} confirm the superiority of the idea of composing AE and sparse AE which leads to improving the performance in terms of accuracy and computational cost. 
	
\section{Conclusion}
\label{sec:conclusion}
In this paper, we have proposed a novel yet efficient semi-supervised method based on AEs which is called AutoIDS. AutoIDS takes advantage of cascading two encoder-decoder neural networks forced to provide a compressed and a sparse representation from normal traffic. These neural networks are known as AE and sparse AE, respectively. In case that these neural networks fail in providing the desire representations for an incoming packet flow, it does not follow the concept of normal traffic. The first detector, i.e., sparse AE, is faster while the second one, i.e., AE, makes a decision about the incoming packet flows more accurately. It is worth mentioning that, a large number of incoming packet flows are processed merely by the first detector. AutoIDS has been evaluated comprehensively on the NSL-KDD dataset. Results confirm the superiority of our method in comparison with the other state-of-the-art approaches.

\ifCLASSOPTIONcaptionsoff
  \newpage
\fi

%\begin{thebibliography}{1}

%\end{thebibliography}
%\end{IEEEbiography}

\ifCLASSOPTIONcaptionsoff
    \newpage
\fi
	
\nocite{*}
\bibliographystyle{IEEEtran}
\bibliography{AutoIDS} 

\end{document}